# Sinhala Language Corpora and Stopwords from a Decade of Sri Lankan Facebook

Yudhanjaya Wijeratne[†], Nisansa de Silva[‡]
† LIRNEasia, 12 Balcombe Place, Colombo, Sri Lanka (yudhanjaya@lirneasia.net)
‡ University of Oregon, 1585 E 13th Ave, Eugene, OR 97403, United States (nisansa@cs.uoregon.edu)

LIRNEasia is a pro-poor, pro-market think tank whose mission is *catalyzing policy change through research to improve people's lives in the emerging Asia Pacific by facilitating their use of hard and soft infrastructures through the use of knowledge, information and technology.* This work was carried out with the aid of a grant from the International Development Research Centre (IDRC), Ottawa, Canada.

# Abstract

This paper presents two colloquial Sinhala language corpora from the language efforts of the *Data, Analysis and Policy* team of LIRNE*asia*, as well as a list of algorithmically derived stopwords. The larger of the two corpora spans 2010 to 2020 and contains 28,825,820 to 29,549,672 words of multilingual text posted by 533 Sri Lankan Facebook pages, including politics, media, celebrities, and other categories; the smaller corpus amounts to 5,402,76 words of only Sinhala text extracted from the larger. Both corpora have markers for their date of creation, page of origin, and content type.

# Introduction

*'The limits of my language mean the limits of my world.'*
– Ludwig Wittgenstein

Sinhala, as with many other languages in the Global South, currently suffers from a phenomenon know as resource poverty [1]. To wit, many of the fundamental tools that are required for easy and efficient natural language analysis are unavailable; many of the more computational components taken for granted in languages like English are either as yet unbuilt, in a nascent stage, and in other cases, lost or retained among select institutions [2].

Adding to the complexity is the fact that Sinhala exhibits diglossia, as noted by Gair [3], and this diglossia extends even unto the script and symbols therein: Silva and Kariyawasam [4] note the existence of Suddha Sinhala, a 'core set' capable of representing the basic sounds of the Sinhala vocabulary, as well as 'misra' or Mixed Sinhalese devised as an adaptation to influences from Pali and Sankrit; and also that in later ages the alphabet was influenced again by English, adding sounds not found in the former sets, like the sound of the English letter 'f'. When combined with the later study of code-mixing (and the effect of English on Sinhala usage and vice versa) by Senaratne [5], the literature suggests a language with multiple, stable forms of usage, both formal and informal varieties, in a process of constant evolution. To engage in the vernacular, corpora reflecting this evolution are essential to enrich the foundations of language research in Sinhala [1].

# Methodology

The Facebook Crowdtangle[1] platform provides default lists - collections of pages put forth by Facebook as belonging to the information ecosystem of a particular country. The list for Sri Lanka contains the Facebook pages of national politicians and political groups; major news and media channels; the pages of major Sri Lankan celebrities; and sports channels. To maximize the amount of Sinhala content available, we used a truncated version of this list, and from them downloaded a total of 1,820,930 posts created between 01-01-2010 and 02-02-2020, thus representing a decade of content poured out into public view on Facebook[2].

[1] https://help.crowdtangle.com/en/articles/4201940-about-us

This data contains one notable defect. Characters that sport the rakārānsaya, such as the word ශ්‍රී- as in “Sri Lanka” - appears as ශ්රී. This may be because of known issues with unicode: the rakārāngśaya, yanśaya and répaya are Sinhala characters that does not exist in the Unicode 13 specification [6][3]. This lack has been noted by Dias [7] as far back as 2010. Simply copying and pasting the ශ්‍රී character on Facebook immediately results on it being decomposed into ශ්රී[4]. What is ultimately recorded in the data is phonetically similar, but inaccurate spelling[5].

## Corpus-Alpha

After stripping the data of irrelevant fields to minimize processing time, the remaining data retained the following attributes:

- **Page Name** - the origin of the post.
- **Create** - this being the timestamp of creation, reduced to a date field in DD-MM-YY format.
- **Type** - this being internal categories that Facebook assigns, based on the content of the post.
- **Message** - the actual text content that the user typed in the post.

The categories present in the Type column are as follows: YouTube, Status, Link, Photo, Native Video, Video, Vine, Live Video Complete, and Live Video.

After stripping out entries that had no informational value in the Message column, we were left with 1,500,917 posts from 533 pages, a mixture of Sinhala, Tamil and English content - all three major languages spoken in Sri Lanka. We consider words to be separated by one or more spaces[6]; by this count there are 29,549,672 words in this dataset.

An examination of quantiles reveals that the median status stands at 13 words and the maximum is of 232 words in length as shown in Table 1.

This set of figures, however, is imprecise. Our goal with this corpus is to capture as much information as possible, including the different vagaries of colloquial spelling on social media and content-sharing habits: therefore, this data includes URLs.

[2] It should be noted that as of the time of writing, the Crowdtangle platform, either by design or by technical limitation, returns only around 200,000 posts per request, or 200 megabytes of data; this number varies, suggesting that it is not a hardcoded cap. Thus this data fetch was performed year-by-year; a single pull for the date range specified resulted in far less data. Likewise, a more granular set of requests may yield more data. The mechanism for sampling is unknown to us: we must therefore assume that it is reasonably representative of the full river of content that Facebook hosts.

[3] For quick reference, see `https://unicode.org/charts/PDF/U0D80.pdf`

[4] Indeed, the same inaccuracy occurred in the Overleaf editor in which we wrote this text.

[5] While pages beginning with ශ්‍රී exist on Facebook, they can be retrieved by a text search for ශ්රී, leaving us to surmise that the rendering hack proposed by Dias [7] in 2010 is still in play a decade later.

[6] As matched by the regular expression \s+

| 0% | 25% | 50% | 75% | 100% |
|---|---|---|---|---|
| 1 | 8 | 13 | 22 | 232 |

Table 1: Distribution of wordcounts at various quantiles.

When cleaned of URLs[7], the result stands at 28,825,820 words. The distribution of wordcounts at various quantiles remains the same.

Due to how search operations and rulesets interact with multilingual text data, we have footnoted some examples observed in practice[8] and chosen to minimize cleaning operations here. Therefore, we can say that this corpus contains between 29,549,672 and 28,825,820 words. Media pages are key contributors of data, as shown in Table 2.

| Contributing page | Number of posts |
|---|---|
| Newsfirst.lk | 99295 |
| HIRU FM | 77786 |
| Sun FM | 75721 |
| Shaa FM | 66761 |
| Tharunayaweb | 62587 |

Table 2: Top five contributors to the corpus.

To better differentiate between the languages contained herein, and to more efficiently extract data, each item of data was annotated for language using fastText library for text classification and representation [8]. fastText provides neural networks trained on language classification tasks; we specifically used the 176.ftz compressed language identification model [9]. This model, with the $k$ parameter set to 2, generates an annotation of the top two most-represented languages in a piece of text; this variable is appended as *langprediction* for each post.

It should be noted that the fastText annotation is not perfect; it was noted that its base training data[9] is unable to account for posts that exhibit codemixing, and thus it erronously assigns false language classifications to said posts[10]. It is this classification error that makes it easier to search for and extract codemixed language from this dataset; otherwise the search would be manual and exhausting[11].

We present this initial dataset as **Corpus-Alpha**.

[7]For cases detected within the dataset by random sampling; there is a non-zero probability of minor artifacts of truncated URLs components still existing, due to the diversity of URL types in social media

[8]For example, the line *Live at 12 News Update:(26 December 2010)* would benefit from the non-alphanumeric characters replaced with spaces, as presently *Update:(26* registers as one word. However, the same operation would turn *it's* into *it* and *s*, thus presenting as two words. Likewise, an operation that searches for and removes text prefixing and suffixing */pages/* - an artefact of Facebook page URLs - will also remove useful information from a post such as *"Do you often interact with new profiles/pages/groups on Facebook?"*

[9]A combination of Wikipedia, Tatoeba and SETimes.

[10]For example, *Maxxxxxxxxxxxxxxxxxxxxxa*, Sri Lankan street slang that borrows heavily from English, is represented using English characters; this it chose to annotate as English and Catalan / Valencian. Likewise, *"Mama Marunath Janathawa Samagai" - Johnston* - a Sinhala statement written in English script - is annotated as English and Indonesian.

[11]Its failings highlight other interesting behaviors. There are 0 posts with Sinhala and Tamil tags appearing together: while a handful of posts do contain Sinhala and Tamil together, their Tamil content is so small that fastText apparently cannot detect it. A qualitative analysis using a random sample of posts tagged as having Tamil content reveals that in this dataset, Sinhala and Tamil are indeed rarely written together. However, a select few words do find themselves neighboring each other in a statistically insignificant number of posts.

## Corpus-Sinhala-Redux

Because Corpus-Alpha is mixed, it may not be desirous for language-specific applications - especially Sinhala, which, of the three languages - English, Sinhala and Tamil - is the most resource-poor. Therefore, we derived from Corpus-Alpha a smaller corpus containing only Sinhala text.

For this we first extracted from Corpus-Alpha the posts tagged for Sinhala by fastText, on the logic that for fastText to detect Sinhala, therein a substantive percentage of a given post must be of Sinhala text. Out of posts with 771,075 posts tagged for Sinhala, 87,922 also have tags for English; this number rises to 208,702 if we add posts tagged for Sinhala and Latin. Thus, in total, 27% of Sinhala posts contain Latinate characters, be they artifacts of codemixing or URLs pasted in the status itself.

From these posts, we removed punctuation[12], URLs, and platform artefacts such as *'featureyoutube', 'indexph', 'and pid and vid and page'*. The pages from *Rivira* and *Sirasa Lakshapathi* - one a news/media platform, the other a program series produced by Sirasa TV - were removed altogether, as status text from them contained a high incidence of URL formatting quirks (*sundayrividaharahtml, editionbreakingnewshtml* and periodical 'click here to view' text that appeared to have been added manually to increase audience interaction with posts. All remaining text was subjected to a purge of all remaining Latinate, Tamil and Chinese characters[13]. Special characters were also removed[14]. This yields a corpus of 364,402 posts, containing a combined 5,402,760 words of status text in the Message column. We show the statistics of this data set in Table 3.

| 0% | 25% | 50% | 75% | 100% |
|---|---|---|---|---|
| 1 | 6 | 9 | 15 | 198 |

Table 3: Distribution of wordcounts at various quantiles.

The number of contributing pages has reduced to 420 (see Appendix B). The top contributors of text are still media organizations that we saw in Table 2, but their ranks have shifted slightly as we show in Table 4.

The standard bag-of-words model describes the frequencies of occurrence of individual words in a given text and thus, generating such a collection is an extremely common procedure in natural language processing, particularly for language classification and modelling tasks. We use it here as a descriptive tool. However, single words alone destroy basic semantic relationships, such as the "bill gates" or "white house" examples used by Bekkerman and Allan [10]. We therefore compute word pairs for Corpus-Sinhala-Redux in order to better display these relationships.

[12]Sinhala, unlike English, does not use inverted commas in the middle of words (ie: it's); pronouns are handled by the ගේ suffix; thus such characters can be safely deleted without splitting a word into two.

[13]As we noted before, fastText's detection is not perfect.

[14]In the process of cleaning, curious interactions were observed between the [:punct:] range in R and Sinhala diacritics such as ්. Notably, the hal kirīma diacritic was removed, presumably as punctuation, while the compound kombuva saha halkirīma (ේ )was preserved. There are various possible explanations for this, from poor support in Unicode for certain Sinhala conventions (as noted earlier) or possible range conflicts between what R considers to be emoji blocs (which are scattered throughout the unicode spec) and Sinhala characters. It was further observed that removing periods individually in R also led to the removal of certain base characters, leaving behind artifacts such as ාතික. Not wishing to take apart R to see what the issue might be, we switched to Perl, using repeated unigram calculations to identify and remove specific unwanted characters, particularly ellipticals, emojis, and other ideograms, on an individual basis.

| Contributing page | Number of posts |
|---|---|
| Ada Derana Sinhala | 44756 |
| Neth FM | 36362 |
| Newsfirstlk | 15769 |
| Hiru News | 12549 |
| Tharunayaweb | 12430 |
| Live at 8 | 10624 |
| Shaa FM | 9897 |
| Hiru Gossip | 8600 |
| BBC News සිංහල | 8589 |
| HIRU FM | 8426 |

Table 4: Top ten contributors to the corpus.

Computing word frequencies shows that Corpus-Sinhala-Redux contains 228,533 unique words and 1,868,589 unique word-pairs. The most common occurrences are shown in Table 5.

| word | freq | word pair | freq |
|---|---|---|---|
| ශ්රී | 26347 | ශ්රී ලංකා | 12514 |
| මහතා | 20961 | වැඩි විස්තර | 6056 |
| දින | 17841 | රාත්රී ට | 5807 |
| ජාතික | 17390 | එක්සත් ජාතික | 5268 |
| රන | 16997 | පාර්ලිමේන්තු මන්ත්රී | 4428 |
| අමාත්ය | 16512 | අද දින | 4396 |
| ඇති | 15249 | මහින්ද රාජපක්ෂ | 4229 |
| ලංකා | 15189 | ගෝඨාභය රාජපක්ෂ | 4196 |
| ගැන | 15031 | වීඩියෝව මෙතැනින් | 4040 |
| සඳහා | 14636 | ලංකා නිදහස | 3346 |

Table 5: Top ten words and word pairs by frequency

It can be readily inferred from Table 5 that the data herein contains a heavy bias towards political conversation, especially since the top word pairs refer to parliamentary ministers, the starting two words of the United National Party (in Sinhala), the starting two words of the Sri Lanka Freedom Party, and two Presidents of Sri Lanka by name. The reason for the frequency counts for the first two words of a political party appear in the high frequency list while the pair made up by the second and third word does notis due to the inflected nature of Sinhala. Depending on the case that is being used, the third word,පක්ෂය gets morphed into පක්ෂයේ, පක්ෂයෙන්, පක්ෂයට and other relevant morphological forms along with the root form පක්ෂය. This results in the dilution of the pairs created by the second and the third word of the party name.

We present this cleaned dataset as **Corpus-Sinhala-Redux**.

## Deriving stopwords from Corpus-Sinhala-Redux

It is by now an axiom of natural language processing that every corpus of text contains words that occur in virtually every sentence, and thus have low informational value [11]; Francis et al. [12] posited that in English, the most common words accounted for some 20-30% of a document. These words are commonly referred to as stopwords, and it is customary to remove them from analysis in most applications, especially those that rely on bag-of-words models, such as most popular topic modelling approaches. English has robust lists of stopwords; Tamil less so; and Sinhala least of all. Therefore it behooves us to extract stopwords from the Corpus-Sinhala-Redux.

However, despite their ubiquity, building a list of stopwords is still a contentious task. Most practitioners favor manually-constructed, domain-specific lists, which take extraordinary time and effort to build (especially from corpora as large as ours) and may not ultimately be worth the effort in the first place [13]. Various algorithmic approaches, most based on the frequency of unigrams in a text, seem to generate usable results in English, though few or none can be said to be demonstrably superior [14]. For some non-English languages, such as Arabic, stopwords have been built with tightly-defined rulesets [15], while in languages closer related to Sinhala, like Sanskrit, a combination of word frequency and manual vetting [16] has yielded suitable results.

Given the proximity of Sinhala to Sanskrit, we too utilized a frequency based method. The most important question in an automated stop word extraction method is the frequency threshold; given that our study is mainly data driven, we opted to let the threshold also be dynamically decided from the data itself.

As such, the first step is to collect the frequencies of words across the entire corpus, as done previously; then, outliers on the lowest end of the distribution (frequency $= 1$) are eliminated[15]. Next, we calculated the standard deviation ($\sigma$) and the mean ($\mu$) of the word frequencies. Using these corpus statistics, then we can standardize the frequencies using the equation 1 where $z$ is the standard score, and $x$ is the word frequency.

$$z = \frac{x - \mu}{\sigma} \tag{1}$$

Next we calculated $-1.5 < Z < 1.5$ for a 93.3% threshold, as shown in the higlighted area under the curve in Fig. 1.

From this process, we obtained the following world list: [ ශ්‍රී ] [ මහතා ] [ දින ] [ ජාතික ] [ කරන ] [ අමාත්‍ය ] [ ඇති ] [ ලංකා ] [ ගැන ] [ සඳහා ] [ කටයුතු ] [ කිරීම ] [ රාජපක්ෂ ] [ වෙනුවෙන් ] [ ගරු ] [ ලෙස ] [ අතර ] [ සිට ] [ ජනාධිපති ] [ විසින් ] [ වැඩි ] [ මෙම ] [ ජනපති ] [ සමග ] [ බලන්න ] [ හිටපු ] [ කිරීමට ] [ විස්තර ] [ පාර්ලිමේන්තු ] [ මන්ත්‍රී ] [ සදහා ] [ මහින්ද ] [ ලබා ] [ අපි ] [ සිදු ] [ නිදහස් ] [ එක්සත් ] [ කරයි ] [ ජනතාව ] [ සංවර්ධන ] [ කරන්න ] [ විය ] [ මහා ] [ රාත්‍රී ] [ ප්‍රධානත්වයෙන් ] [ පක්ෂ ] [ අවස්ථාව ] [ පත් ] [ රාජ්‍ය ] [ කොළඹ ] [ පැවති ] [ වැඩ ] [ මාධ්‍ය ] [ ජනතා ] [ කිරීමේ ].

Translated to English, this would be: [ sri ] [ mister ] [ day ] [ national ] [ do ] [ minister ] [ enough ] [ lanka ] [ about ] [ for ] [ affairs ] [ doing ] [ rajapaksa ] [ for ] [ the honorable ] [ as the ] [ while ] [ from ] [ president ] [ by ] [ higher ] [ this ] [ president (*alternate/poetic spelling*)] [ with a ] [ see ] [ former ] [ to ] [ details ] [ parliamentary ] [ member (*of legislative body*) ] [ for (*in this case, a colloquial misspelling of the more formal variant above*)] [ mahinda ] [ obtaining ] [ we ] [ done ] [ free ] [ united ] [ does ] [ people ] [ development ] [ do (*conjugation*)] [ it was ] [ great ] [ night ] [ chiefly ] [ parties ] [ the situation ] [ appoint ] [ of the state ] [ colombo ] [ existing ] [ work ] [ the media ] [ people ] [ doing ].

[15] In this study we have been very lenient in the definition of outliers to avoid data loss.

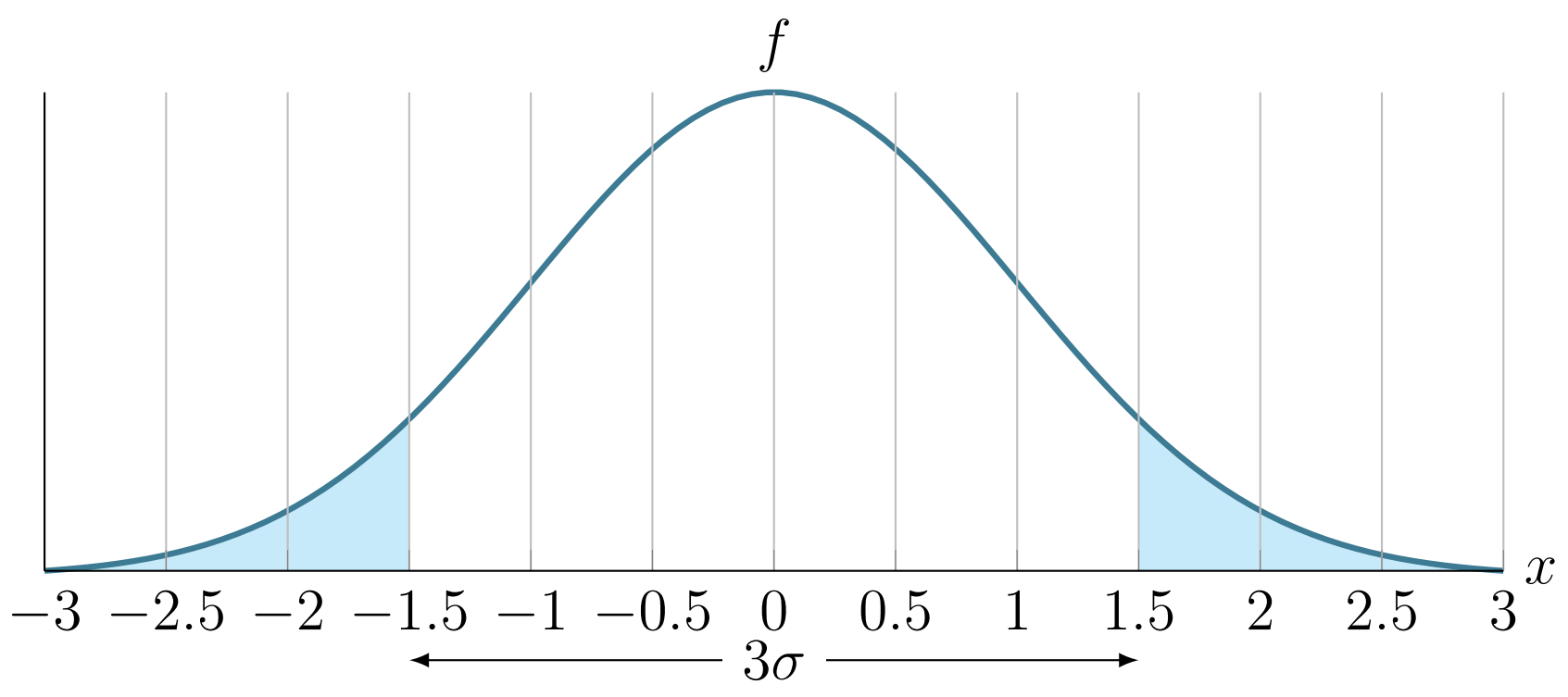

Figure 1: Statistical range of stop words at $-1.5 < Z < 1.5$

There are both similarities and dissimilarities to commonly used English stopwords lists, such as those in Python's NLTK package[16] and R's stopwords package[17].

We present this list as a set of stopwords derived from **Corpus-Sinhala-Redux**.

## Related work

Upeksha et al. [17] created an ambitious 119-million-word corpus project (in which a co-author of this paper took part). Though the portal links referenced in official documentation lead to 404s [18], a copy is preserved in the archives of one of the co-authors of this paper hosted at Open Science Framework (OSF)[18]. Notably, Upeksha et al. drew from newspapers, text books, wikipedia, the Mahavamsa, fiction, blogs, magazines and gazettes, all of which require a degree of formal Sinhala; we present our work, which uses social media data, as a colloquial complement to this dataset.

## Conclusion

In this paper we present two Sinhala corpora and a list of stopwords, both extracted from Facebook, which is a rich source of colloquial text data and invaluable for studying resource-poor languages. Corpus-Alpha is extracted from Facebook pages for a period from 2010 to 2020, and consists of between 28,825,820 to 29,549,672 words of text posted by a variety of Sri Lankan Facebook pages. Corpus-Sinhala-Redux, which is derived from Corpus-Alpha, contains 5,402,760 words of only Sinhala text from 420 of those pages.

[16] https://www.nltk.org/nltk_data
[17] https://cran.r-project.org/web/packages/stopwords/stopwords.pdf
[18] https://osf.io/a5quv/files/

The objective of Corpus-Alpha is to capture as much information as possible, and thus present a rich source for discourse analysis and codemixing between the three languages in use in Sri Lanka, with a bias towards Sinhala. It contains text in English, Sinhala and Tamil, all three major languages spoken in Sri Lanka; additionally, it contains punctuation, URLs, ideograms such as emojis, and serves as a snapshot of Sri Lankan discourse on Facebook[19].

The objective of Corpus-Sinhala-Redux is to provide a collection of colloquial Sinhala commonly used on social media, more immediately suited to Sinhala-specific language applications. Corpus-Sinhala-Redux showcases specific artifacts brought about by how Facebook records Sinhala text, but has been cleaned of ideograms, punctuation, and other symbols not directly relevant to this process. A list of stopwords have been algorithmically derived from Corpus-Sinhala-Redux.

In this paper we have documented both the processes involved in creating these corpora and various platform and language-related nuances around the topic. Both corpora and stopwords list are made available under open access terms at https://github.com/LIRNEasia/FacebookDecadeCorpora. It is our hope that this data will be used to improve natural language processing applications in Sinhala.

## Acknowledgements

This research has been made possible through a grant from the International Development Research Centre, Canada (IDRC) and Facebook's generous access to the Crowdtangle platform.

[19] Given that English occupies a curious role in Sri Lanka - first as the language of colonizers, and today as marker of social class [19], the examination of language in conjuction with Page Names may yield interesting observations on the audience that these pages seek to target.

# Appendix:
# List of highest contributors to corpus-redux-sinhala

| Contributing Page | Number of posts |
|---|---|
| Ada Derana Sinhala | 44756 |
| Neth FM | 36362 |
| Newsfirstlk | 15769 |
| Hiru News | 12549 |
| Tharunayaweb | 12430 |
| Live at 8 | 10624 |
| Shaa FM | 9897 |
| Hiru Gossip | 8600 |
| BBC News | 8589 |
| HIRU FM | 8426 |
| Neth News | 7850 |
| Dinamina | 7278 |
| Mawbima | 6265 |
| Swarnavahini | 6179 |
| Newslk | 5638 |
| Ada | 5041 |
| Sri Lanka Mirror | 4996 |
| Hiru TV | 4419 |
| Siyatha FM | 4269 |
| Lankadeepa | 4125 |
| Neth Gossip | 3863 |
| Ranjan Ramanayake | 3510 |
| Kanaka Herath | 2755 |
| Ada Derana Biz Sinhala | 2531 |
| Wimal weerawansa | 2518 |
| United National Party | 2335 |
| Udaya Prabhath Gammanpila | 2314 |
| Anura Kumara Dissanayake | 2219 |
| FM Derana | 2145 |
| Rajitha Senaratne | 2142 |
| Mirror Arts | 2125 |
| Patali Champika Ranawaka | 2101 |
| Silumina | 2095 |
| Harshana Rajakaruna | 1830 |
| S M Marikkar | 1807 |
| Wajira Abeywardena | 1763 |
| JVP Srilanka | 1624 |
| Resa Newspaper | 1613 |
| Sajith Premadasa | 1595 |
| ENewsFirst | 1569 |